\documentclass{article}
\usepackage{hyperref} %
\usepackage[accepted]{icml2024}
\usepackage{graphicx}

\usepackage{microtype}
\usepackage{amsmath}
\usepackage{amssymb}
\usepackage{amsthm}
\usepackage{booktabs}
\usepackage{caption}
\usepackage{cuted}
\usepackage{mathtools}
\usepackage{subfigure}
\usepackage{xspace}
\usepackage[textsize=tiny]{todonotes}
\usepackage[capitalize,noabbrev]{cleveref}

\makeatletter
\renewcommand{\paragraph}{%
  \@startsection{paragraph}{4}%
  {\z@}{-0.5em}{-0.5em}%
  {\normalfont\normalsize\bfseries}%
}
\makeatother

\newcommand{\method}{IM-3D\xspace}

\theoremstyle{plain}

\theoremstyle{definition}

\theoremstyle{remark}

\icmltitlerunning{\method: Iterative Multiview Diffusion and Reconstruction for High-Quality 3D Generation}

\begin{document}

\twocolumn[
\icmltitle{\method: Iterative Multiview Diffusion and Reconstruction for \\ High-Quality 3D Generation}
\icmlsetsymbol{equal}{*}
\begin{icmlauthorlist}
\icmlauthor{Luke Melas-Kyriazi}{equal,meta,oxford}
\icmlauthor{Iro Laina}{oxford}
\icmlauthor{Christian Rupprecht}{oxford}
\icmlauthor{Natalia Neverova}{meta}
\icmlauthor{Andrea Vedaldi}{meta}
\icmlauthor{Oran Gafni}{meta}
\icmlauthor{Filippos Kokkinos}{equal,meta}
\end{icmlauthorlist}
\vspace{-2em}
\icmlaffiliation{meta}{Meta}
\icmlaffiliation{oxford}{University of Oxford, Oxford, UK}
\icmlcorrespondingauthor{Filippos Kokkinos}{fkokkinos@meta.com}
\icmlcorrespondingauthor{Luke Melas-Kyriazi}{lukemk@robots.ox.ac.uk}
\icmlkeywords{Machine Learning, ICML}
] %

\printAffiliationsAndNotice{\icmlEqualContribution}

\setlength\stripsep{1em}
\begin{strip}
\centering
\includegraphics[width=0.95\textwidth]{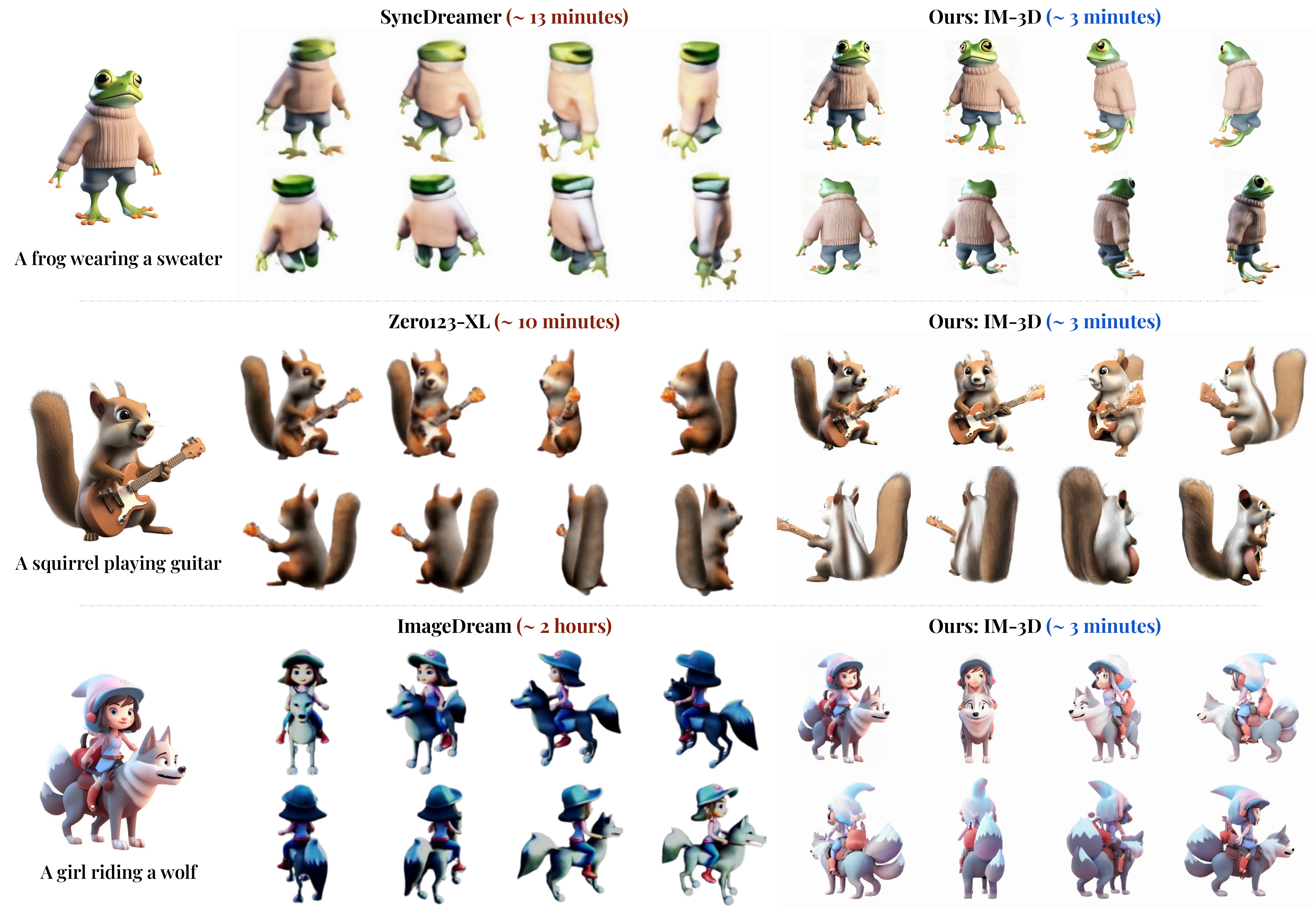}
\captionof{figure}{\method generates high-quality and faithful 3D assets from text/image pair.
It eschews Score Distillation Sampling (SDS) for robust 3D reconstruction of the output of a video diffusion model, tuned to generate a 360\textdegree{} video of the object.\vspace*{7mm}
}\label{fig:splash}
\end{strip}

\begin{center}
\large{\textbf{Abstract}}
\end{center}\vspace*{-10pt}
Most text-to-3D generators build upon off-the-shelf text-to-image models trained on billions of images.
They use variants of Score Distillation Sampling (SDS), which is slow, somewhat unstable, and prone to artifacts.
A mitigation is to fine-tune the 2D generator to be multi-view aware, which can help distillation or can be combined with reconstruction networks to output 3D objects directly.
In this paper, we further explore the design space of text-to-3D models.
We significantly improve multi-view generation by considering video instead of image generators.
Combined with a 3D reconstruction algorithm which, by using Gaussian splatting, can optimize a robust image-based loss, we directly produce high-quality 3D outputs from the generated views.
Our new method, \method, reduces the number of evaluations of the 2D generator network 10-100$\times$, resulting in a much more efficient pipeline, better quality, fewer geometric inconsistencies, and a high yield of usable 3D assets.

\section{Introduction}%
\label{s:intro}

All state-of-the-art open-world text-to-3D generators are built on top of off-the-shelf 2D generators  trained on billions of images.
This is necessary because there isn't enough 3D training data to directly learn generators that can understand language and operate in an open-ended manner.
However, the best way of building such models is still debated.

One approach is to perform 3D distillation by adopting Score Distillation Sampling (SDS)~\cite{poole23dreamfusion:} or one of its variants.
These models can work on top of nearly any modern 2D generator, but they require tens of thousands of evaluations of the 2D generator, and can take hours to generate a single asset.
They are also prone to artifacts and may fail to converge.
Mitigating these shortcomings inspired a significant body of research~\cite{wang23prolificdreamer:}.

The fundamental reason for these limitations is that the underlying 2D generator is not 3D aware.
SDS slowly makes the different views of the 3D object agree with the 2D model, which characterizes them independently of each other.
Several authors~\cite{shi23mvdream:,wang23imagedream:,shi23zero123:} have shown that fine-tuning the 2D generator to understand the correlation between different views of the object significantly facilitates distillation.
More recently, approaches such as~\cite{li23instant3d:} avoid distillation entirely and instead just reconstruct the 3D object from the generated views.
However, in order to compensate for defects in multi-view generation, they must incorporate very large 3D reconstruction networks.
Ultimately, these approaches are many times faster than distillation, but quality is limited.

In this paper, we explore the benefits of further increasing the quality of multi-view generation and how this might affect the design space of future text-to-3D models.
We are inspired by the fact that, in the limit, a 2D generator could output enough consistent views of the object to afford simple multi-view reconstruction, sidestepping distillation and reconstruction networks entirely.

To this end, we introduce \method, a text-to-3D generation approach that leverages \textbf{I}terative \textbf{M}ultiview diffusion and reconstruction (\cref{fig:splash,fig:methods_figure}).
\method is based on significantly boosting the quality of the multi-view generation network
by switching from a text-to-image to a text-to-video generator network.
Specifically, we pick Emu Video~\cite{girdhar23emu-video:}, a video generator conditioned both on a reference image and a textual prompt.
Our first contribution is to show that Emu Video can be fine-tuned, using a relatively small number of synthetic 3D assets, to generate directly up to 16 high-resolution consistent views ($512\times512$) of the object.
While Emu Video is in itself an iterative model based on diffusion,
by adopting a fast sampling algorithm, the views can be generated in a few seconds and in a small number of iterations.

Our second contribution is to show that we can extract a high-quality 3D object by \emph{directly} fitting a 3D model to the resulting views---without distillation or reconstruction networks---quickly and robustly.
To do so, we rely on a 3D reconstruction algorithm based on Gaussian splatting (GS)~\cite{kerbl233d-gaussian}.
The importance of GS is that it affords fast differentiable rendering of the 3D object, which allows the use of image-based losses like LPIPS\@.
The latter is key to bridging the small inconsistencies left by the 2D generator without requiring ad-hoc reconstruction models.

Third, we notice that, while this process results in mostly very good 3D models, some inconsistencies may still remain.
We thus propose to close the loop and feed the 3D reconstruction back to the 2D generator.
In order to do so, we simply render noised images of the 3D object and restart the video diffusion process from those.
This approach is closer in spirit to SDS as it builds consensus progressively, but the feedback loop is closed two or three times per generated asset, instead of tens of thousands of times.

There are many advantages to our approach.
Compared to SDS, it reduces dramatically the number of evaluations of the 2D generator network.
Using a fast sampler, generating the first version of the multi-view images requires only around 40 evaluations.
Iterated generations are much shorter (as they start from a partially denoised result), at most doubling the total number of evaluations.
This is a 10-100$\times$ reduction compared to SDS.
The 3D reconstruction is also very fast, taking only a minute for the first version of the asset, and a few seconds for the second or third.
It also sidesteps typical issues of the SDS such as artifacts (e.g., saturated colors, Janus problem), lack of diversity (by avoiding mode seeking), and low yield (failure to converge).
Compared to methods like~\cite{li23instant3d:}, \method is slower, but achieves much higher quality, and does not require to learn large reconstruction networks, offloading most of the work to 2D generation instead.

In a nutshell, our contribution is to show how video generator networks can improve consistent multi-view generator to a point where it is possible to obtain state-of-the-art and efficient text/image-to-3D results without distillation and without training reconstruction networks.

\section{Related work}%
\label{s:related-work}

\paragraph{3D Distillation.}

3D distillation is the process of extracting a 3D object from a 2D neural network trained to generate images from text, or otherwise match them to text. For example, methods like DreamFields~\cite{jain22zero-shot} do so starting from the CLIP image similarity score.
However, most recent methods build on diffusion-based image generators that utilize variants of the Score Distillation Sampling (SDS) loss introduced with DreamFusion~\cite{poole23dreamfusion:}.
Fantasia 3D~\cite{chen23fantasia3d:} disentangles illumination from materials.
Magic3D~\cite{lin22magic3d:} reconstructs high-resolution texture meshes.
RealFusion~\cite{melas-kyriazi23realfusion} starts from a reference image and fine-tunes the prompt of a 2D generator to match it, distilling a 3D object afterwards.
Make-it-3D~\cite{tang23make-it-3d:} also starts from a 2D image, combining SDS with a CLIP loss with respect to the reference image and a depth prior.
HiFi-123~\cite{yu23hifi-123:} uses DDIM inversion to obtain the code for the reference image.
ATT3D~\cite{lorraine23att3d:} develops an amortized version of SDS\@, where several variants of the same object are distilled in parallel.
HiFA~\cite{zhu23hifa:} reformulates the SDS loss and anneals the diffusion noise.
DreamTime~\cite{huang23dreamtime:} also proposes to optimize the noise schedule.
ProlificDreamer~\cite{wang23prolificdreamer:},
SteinDreamer~\cite{wang23steindreamer:}, Collaborative SDS~\cite{kim23collaborative} and Noise-free SDS~\cite{katzir23noise-free} improve the variance of the SDS gradient estimate.
DreamGaussian~\cite{tang23dreamgaussian:}, GaussianDreamer~\cite{yi23gaussiandreamer:} and~\cite{chen23text-to-3d} apply Gaussian splatting to the SDS loss.

\paragraph{Methods using multi-view generation.}

Many methods have proposed to use multi-view generation to improve 3D generation.
For multi-view generation, the most common approach is Zero-1-to-3~\cite{liu23zero-1-to-3:}, which fine-tunes the Stable Diffusion (SD) model to generate novel views of an object.
Zero123++~\cite{shi23zero123:} further improves on this base model in various ways, including generating directly a grid of several multi-view images. Cascade-Zero123~\cite{chen23cascade-zero123:} proposes to apply two such models in sequence: the first to obtain approximate multiple views of the object, and the second to achieve better quality views conditioned on the approximate ones.

Magic123~\cite{qian23magic123:} and DreamCraft3D~\cite{sun23dreamcraft3d:} combine Zero-1-to-3 and SD\@.
They start from a generated 2D image, extract depth and normals, and apply the RealFusion / DreamBooth technique to fine-tune the 2D diffusion model to generate different views of the object.

MVDream~\cite{shi23mvdream:} directly generates four fixed viewpoints of an object from a text prompt.
Consistent123~\cite{weng23consistent123:} uses a different form of cross-view attention and generates several views sufficient for direct reconstruction.
ConsistNet~\cite{yang23consistnet:} introduces an explicit 3D pooling mechanism to exchange information between views.
ImageDream~\cite{wang23imagedream:} extends MVDream to start from a given input image, and proposes a new variant of image conditioning compared to that of Zero-1-to-3.
RichDreamer~\cite{qiu23richdreamer:} further learns to generate normals and separation between material and lighting.

Viewset Diffusion~\cite{szymanowicz23viewset}, Forward Diffusion~\cite{tewari23diffusion}, SyncDreamer~\cite{liu23syncdreamer:} and DMV3D~\cite{xu2023dmv3d} denoise multiple views of the 3D object simultaneously to improve consistency.

3DGen~\cite{gupta233dgen:} learns a latent space to encode 3D objects using a VAE-like technique. The latent space is then used by a diffusion model that draws samples from it. However, this approach is not very scalable as it requires training the model from scratch using 3D data.
HexaGen3D~\cite{mercier24hexagen3d:} extends 3DGen to use features from an SD model instead, thus increasing the scalability of the approach.
A concurrent work is ViVid-1-to-3~\cite{kwak23vivid-1-to-3:}, which also uses a video generator for multi-view generation, but does not produce any 3D assets (only novel views).

\paragraph{Non-SDS methods.}

Some text-to-3D methods perform ``direct'' 3D reconstruction on top of generated views without using SDS\@.
One--2--3--45~\cite{liu23one-2-3-45:} compensates for the shortcomings of the multi-view generator by training a reconstruction network.
Instant3D~\cite{li23instant3d:} is similar, but based on a much larger reconstruction model~\cite{hong23lrm:}.
Wonder3D~\cite{long23wonder3d:} further learns to generate multiple views of a given input image together with the corresponding normal maps, which are then used to reconstruct the 3D object.
AGG~\cite{xu24agg:} builds a single-image reconstruction network on top of Gaussian splatting.
CAD~\cite{wan23cad:} learns a 3D generator network from image samples using a 2D diffusion model, replacing the SDS loss with adversarial training.

Our approach also eschews the SDS loss, but shows that it is possible to offload most of the modelling burden to the 2D generator network, utilizing a straightforward and efficient 3D reconstruction algorithm.

\section{Method}%
\label{s:method}

\begin{figure*}
\centering
\includegraphics[width=0.87\textwidth]{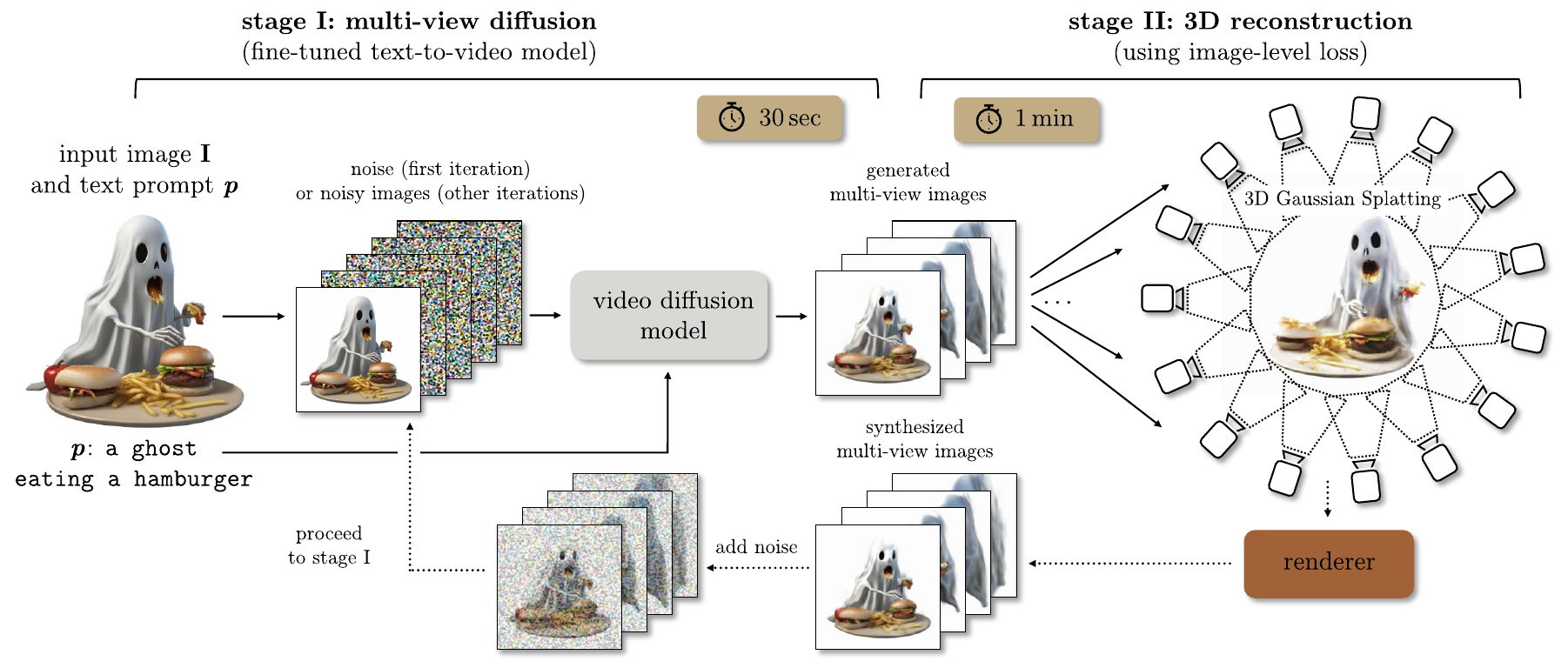}
\caption{\textbf{Overview of \method.}
Our model starts from an input image (e.g., generated from a T2I model).
It feeds the latter into an image-to-video diffusion model to generate a turn-table like video.
The latter is plugged into 3D Gaussian Splatting to \emph{directly} reconstruct the 3D object using image-based losses for robustness.
Optionally, renders of the objects are generated and fed back to the video diffusion model, repeating the process for refinement.}%
\vspace{-3mm}
\label{fig:methods_figure}
\end{figure*}

\newcommand{\I}{\mathbf{I}}
\newcommand{\J}{\mathbf{J}}
\newcommand{\p}{\boldsymbol{p}}

We first describe our video-based multi-view generator network in \cref{s:mv-video} and its training data in \cref{s:data}, followed by a description of the robust 3D reconstruction module in \cref{s:fast-reconstruction} and of iterative refinement in \cref{s:iterated-generation}.
{An overview of our method is shown in \Cref{fig:methods_figure}}.

\subsection{Multi-view as video generation}%
\label{s:mv-video}

Our multi-view generation model is based on fine-tuning an existing text-to-video (T2V) generator network Emu~Video~\cite{girdhar23emu-video:}.
First, it utilizes a text-to-image (T2I) model (Emu~\cite{dai23emu:}) to generate an initial image $\I$ corresponding to the given textual prompt $\p$.
Second, the image $\I \in \mathbb{R}^{3\times H\times W}$ and the text prompt $\p$ are fed into a second generator, which produces up to $K=16$ frames of video $\J \in \mathbb{R}^{K\times 3\times H\times W}$, utilizing $\I$ as guidance for the first frame.
Notice that, while the model is trained such that $\I \approx \J_1$, this is not an exact equality.
Instead, the model draws a sample $\J$ from a learned conditional distribution $p(\J|\I,\p)$, which allows it to slightly deviate from the input image to better fit in the generated video.
An advantage of Emu Video compared to other video generators is that the video frames $\J$ are already high-quality and high-resolution, without requiring sophisticated coarse-to-fine sampling schemes. %
It is architectured as a fine-tuned version of the original T2I Emu network with some modifications to account for the temporal dependencies between frames. %

Starting from the pre-trained Emu Video model, we then fine-tune it to generate a particular kind of video, where the camera moves around a given 3D object, effectively generating \emph{simultaneously} several views of it, in a turn-table-like fashion.
In order to do so, we consider an internal dataset of synthetic 3D objects, further described in \cref{s:data}.
This dataset provides us with training videos $\mathcal{J}=\{(\J_n,\I_n,\p_n)\}_{n=1}^N$, each containing $K=16$ views of the object taken at fixed angular interval and a random but fixed elevation, the initial image $\I_n=[\J_n]_1$, and the textual prompt $\p_n$.
The camera distance is fixed across all renders.

Differently from many prior multi-view generation networks, we \emph{do not} pass the camera parameters to the model; instead, we use a fixed camera distance and orientation, randomizing only the elevation.
The model simply learns to produce a set of views that follow this distribution.

Like most image and video generators, Emu Video is based on \emph{diffusion} and implements a denoising neural network
$
\hat \epsilon(\J_t, t, \I, \p)
$
that takes as input a noised video
$
\J_t = \sqrt{1 - \sigma_t^2} \J + \sigma_t \epsilon,
$
where $\epsilon \sim N(0,I)$ is Gaussian noise and $\sigma_t \in [0,1]$ is the noise level, and tries to estimate the noise $\epsilon$ from it.
The training uses the standard diffusion loss
$
\mathcal{L}_\text{diff}(\hat\epsilon|\J,\I,\p, t, \epsilon)
=
w^\text{diff}_t \cdot \|\hat \epsilon(\J_t, t, \I, \p) - \epsilon \|^2
$
where $(\J,\I,\p)\in\mathcal{J}$ is a training video, $\epsilon$ is a Gaussian sample, $t$ is a time step, also randomly sampled, and $w_t$ is a corresponding weighing factor.
To finetune Emu Video, we use $\mathcal{L}_\text{diff}$, but freeze all parameters except for the temporal convolutional and attention layers.

\subsection{Data}%
\label{s:data}

The dataset $\mathcal{J}$ used to train our model consists of turn-table-like videos of synthetic 3D objects.
Several related papers in multi-view generation also use synthetic data, taking Objaverse~\cite{deitke22objaverse:} or Objaverse-XL~\cite{deitke23objaverse-xl:} as a source.
Here, we utilize an in-house collection of 3D assets of comparable quality, for which we generate textual descriptions using an image captioning network.

Similar to prior works~\cite{li23instant3d:}, we use a subset of 100k assets selected for quality, as determined by the CLIP~\cite{radford21learning} alignment between rendered images and textual descriptions.
Each video $\J \in \mathcal{J}$ is obtained by sampling one of the 100k assets, choosing a random elevation in $[0,\pi/4]$, and then placing the camera around the object at uniform ($2\pi / K$ degree) intervals.

\subsection{Fast and robust reconstruction}%
\label{s:fast-reconstruction}

To generate a 3D asset from a prompt $\p$, we first sample an image $\I \sim p(\I|\p)$ from the Emu image model, followed by sampling a multi-view video $\J \sim p(\J|\I,\p)$ from the fine-tuned Emu Video model.
Given the video $\J$, we then \emph{directly} fit a 3D model $G$.
While there are many possible choices for this model, here we use Gaussian splatting~\cite{kerbl233d-gaussian}, a radiance field that uses a large number of 3D Gaussians to approximate the 3D opacity and color functions.

Given the 3D model $G$ and a camera viewpoint $\Pi$, the \emph{differentiable} Gaussian splatting renderer produces an image $\hat \I = \mathcal{R}(G, \Pi)$.
Compared to other methods such as NeRF~\cite{mildenhall20nerf:}, or even faster versions such as DVGO~\cite{sun22direct} or TensoRF~\cite{chen22tensorf:}, the key advantage of Gaussian splatting is the efficiency of the differentiable renderer, both in time and space, which allows rendering a \emph{full} high-resolution image $\I$ at each training iteration instead of just selected pixels as in most prior works.
Because of this fact, we can utilize \emph{image-level} losses such as LPIPS~\cite{zhang18the-unreasonable}, i.e.,
$
\mathcal{L}_\text{LPIPS}(\hat\I, \I)
=
\sum_{q=1}^Q
\|
w_q \odot (\Phi_q(\hat\I) - \Phi_q(\I))
\|^2
$
where $\Phi_q : \mathbb{R}^{3\times H\times W} \rightarrow \mathbb{R}^C$ is a family of $Q$ patch-wise feature extractors implemented by the VGG-VD neural network~\cite{simonyan15very}.
We also utilize a second image-based loss $\mathcal{L}_\text{MS-SSIM}$, the multi-scale structural similarity index measure (MS-SSIM)~\cite{ms_ssim}. Finally, we use a mask loss $\mathcal{L}_\text{Mask}$ with masks obtained using the method introduced in~\cite{qin2022}.
In our ablation studies, we show the significant benefits of using these image-based losses rather than the standard pixel-wise RGB loss $\mathcal{L}_\text{RGB}(\hat\I, \I) = \|\hat\I - \I \|^2$.
Our final loss is the weighted loss combination
$
\mathcal{L} =
w_\text{LPIPS} \mathcal{L}_\text{LPIPS} +
w_\text{SSIM} \mathcal{L}_\text{SSIM} + w_\text{Mask} \mathcal{L}_\text{Mask}  .
$
The object $G$ is thus reconstructed via direct optimization, i.e.,
$
G^*
=
\operatornamewithlimits{argmin}_{G} %
\sum_{k=1}^K
\mathcal{L}(
  \mathcal{R}(G, \Pi_k),
  [\J]_k
)
$
where $[\J]_k$ denotes the $k$-th image in the video.

\begin{table*}[t]
\centering
\caption{\textbf{Faithfulness to the textual and visual prompts of image sequences synthesised or rendered by various methods}.  Assessed on the prompt list from~\cite{wang23imagedream:,shi23mvdream:}.}%
\label{tab:clip_scores}
\vspace{-1mm}
\begin{tabular}{lc|cc|cc}
\toprule
 &  & \multicolumn{2}{c|}{synthesized view} & \multicolumn{2}{c}{re-rendered view}  \\ \midrule
\multicolumn{1}{l|}{model} & \multicolumn{1}{c}{\begin{tabular}[c]{@{}c@{}}Time (min)\!\!\!\end{tabular}} & CLIP (Text) & CLIP (Image) & CLIP (Text) & CLIP (Image) \\ \midrule
\multicolumn{1}{l|}{\textit{SDXL~\cite{podell23sdxl:} [upper bound]}} & \textit{0.03} & \textit{33.33}  & \textit{100}  & --- & ---   \\ \midrule
\multicolumn{1}{l|}{MVDream~\cite{shi23mvdream:}} & 72 & 31.26 \textpm{2.9} & 76.44 \textpm{6.5}& 30.63 \textpm{2.7} & 76.94 \textpm{5.2} \\
\multicolumn{1}{l|}{Zero123XL~\cite{deitke23objaverse-xl:}} & 10 & 19.58 \textpm{1.3} & 60.29 \textpm{5.8} & 29.06 \textpm{3.3} & 81.33 \textpm{6.9} \\
\multicolumn{1}{l|}{Magic123~\cite{qian23magic123:}} & 15 & --- & --- & 29.51 \textpm{4.7}  & 84.14 \textpm{10.2} \\
\multicolumn{1}{l|}{SyncDreamer~\cite{liu23syncdreamer:}} & 13 & 27.76 \textpm{3.0} & 77.26 \textpm{7.2} & 26.22 \textpm{3.4} & 74.95 \textpm{6.6} \\
\multicolumn{1}{l|}{ImageDream~\cite{wang23imagedream:}} & 120 & 31.08 \textpm{3.4} & 85.39 \textpm{5.8} & 30.73 \textpm{2.3} & 83.77 \textpm{5.2}  \\
\multicolumn{1}{l|}{OpenLRM~\cite{hong23lrm:}} & \textbf{0.17} & --- & --- & 29.75 \textpm{3.2} & 83.08 \textpm{9.5} \\
\multicolumn{1}{l|}{One2345++~\cite{liu23one-2-3-45:}} & 0.75 & --- & --- & 29.71 \textpm{2.3} & 83.78 \textpm{6.4} \\
\multicolumn{1}{l|}{{\method (ours)}} & 3 & \textbf{31.92} \textpm{1.6} & \textbf{92.38} \textpm{5.1} & \textbf{31.66} \textpm{1.7} & \textbf{91.40} \textpm{5.5} \\
\bottomrule
\end{tabular}
\vspace{-1mm}
\end{table*}

\vspace{-2mm}
\subsection{Fast sampling and iterative generation}%
\label{s:iterated-generation}

The SDS loss can be seen as a way to bridge the gap between image generators that are unaware of 3D objects and their 3D reconstructions, absorbing multi-view consistency defects in the generation.
Because our model is rather view-consistent from the outset, and because we can use robust reconstruction losses, the SDS loss is unnecessary.
Instead, given a prompt $\p$, we simply generate an image $\I$, followed by video $\J$, and then fit a 3D object $G$ to the latter.

One main advantage is that this \emph{dramatically} reduces the number of model evaluations compared to using the SDS loss.
Optimizing the SDS loss is (approximately) the same as ascending the score, i.e., the gradient
$
\nabla \log p(\J_t|\I,\p)
$
of the log distribution over noised videos (or images), so the optimization of the asset $G$ can be seen as a form of multi-view mode seeking.
The score is obtained from the same network $\hat\epsilon$. %
However, despite the conditioning on a specific textual prompt $\p$ and input view $\I$, the sampled distribution is rather wide, requiring a very large number (thousands) of iterations to converge to a mode; furthermore, regressing to a mode reduces the diversity and quality of the output.

In our case, the network $\hat \epsilon$ is used to generate directly a \emph{single} video $\J$, which is then reconstructed without further invocations to the model.
Because the video $\J$ is already sufficiently view-consistent, the 3D reconstruction converges quickly to a good solution.
Furthermore, we can adopt fast stochastic ODE solvers such as DPM++~\cite{lu22dpm-solver:} to further reduce the number of model evaluations to obtain the video in the first place.
Overall, compared to using the SDS loss, the number of model evaluations is reduced by a factor $10$-$100\times$ (see the Appendix for additional analysis). %

Despite the overall consistency of generated videos $\J$, they are still not perfect.
We thus additionally compensate for such inconsistencies during model fitting, but still without resorting to the SDS loss.
Instead, we alternate 3D reconstruction and video generation.
To do so, once the first video $\J$ and corresponding 3D model $G^*$ are obtained, we use the latter to generate a video $\J^* = \mathcal{R}(G^*, \Pi)$ using the 3D renderer, sample an intermediated noised video $\J^*_t$ by adding noise to it as shown above, and then invoking the video generator again to obtain a denoised and updated video $\J'$.

We iterate this process two times.
This is vastly faster than using the SDS loss while still being highly robust.  %

\begin{figure*}
\centering
\includegraphics[width=0.99\textwidth]{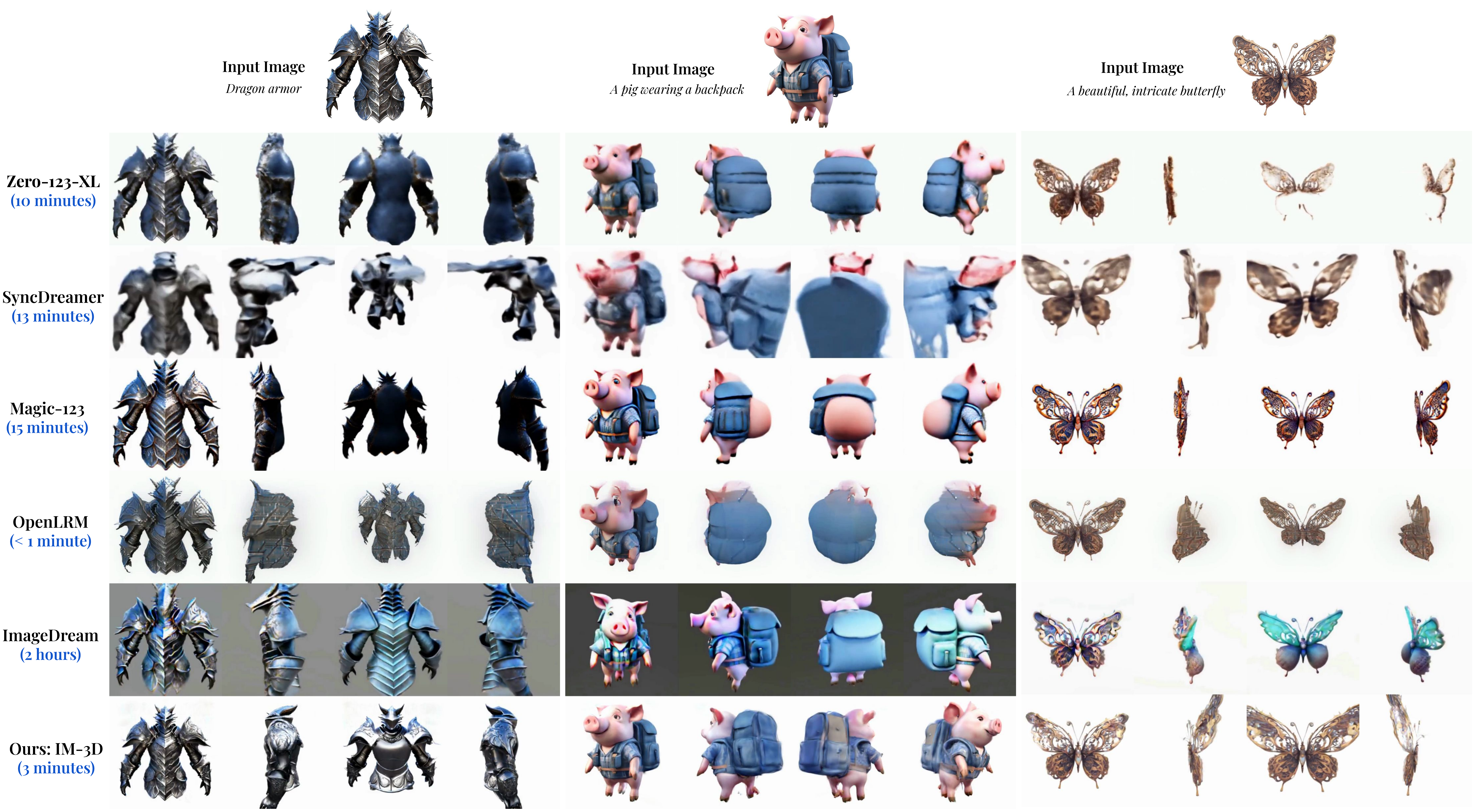}
\caption{\textbf{Qualitative Comparisons.} Our method \method (last row) and others for the same text/image prompt pairs.
For \method, we show the final GS reconstruction (which guarantees multi-view consistency).
We match the input image faithfully and obtain high-quality, detailed reconstructions in just 3 minutes.
Faster methods such as OpenLRM are also much worse.
}
\label{fig:compairson_figure}
\end{figure*}
\begin{figure*}
\centering
\includegraphics[width=0.99\textwidth]{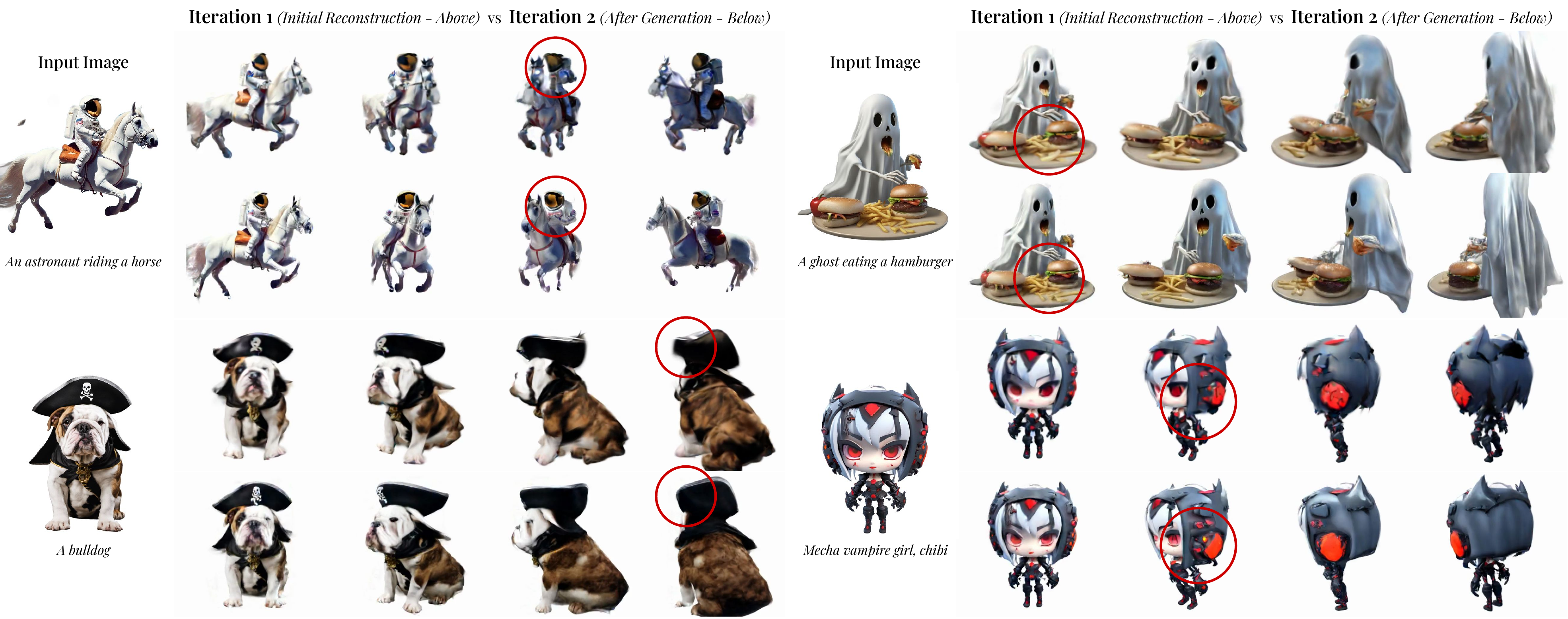}
\caption{\textbf{A visualization of reconstruction quality over multiple iterations of multiview diffusion and reconstruction.} 
We compare the initial reconstructions obtained by our model (i.e. the result of training on our initial generated videos) to the result after one iteration of reconstruction and refinement. 
We see that although the initial reconstructions have reasonable shapes, they lack fine-grained details due to small inconsistencies in the generated multiview images. 
After one iteration of noising, denoising, and reconstruction, our method resolves these inconsistencies and produces 3D assets with significantly higher levels of detail (as highlighted by the red circles above). 
}%
\label{fig:iterations_figure}
\end{figure*}
\begin{figure}[t]
\centering
\vspace{-3mm}
\hspace{-4mm}
\includegraphics[width=0.48\textwidth]{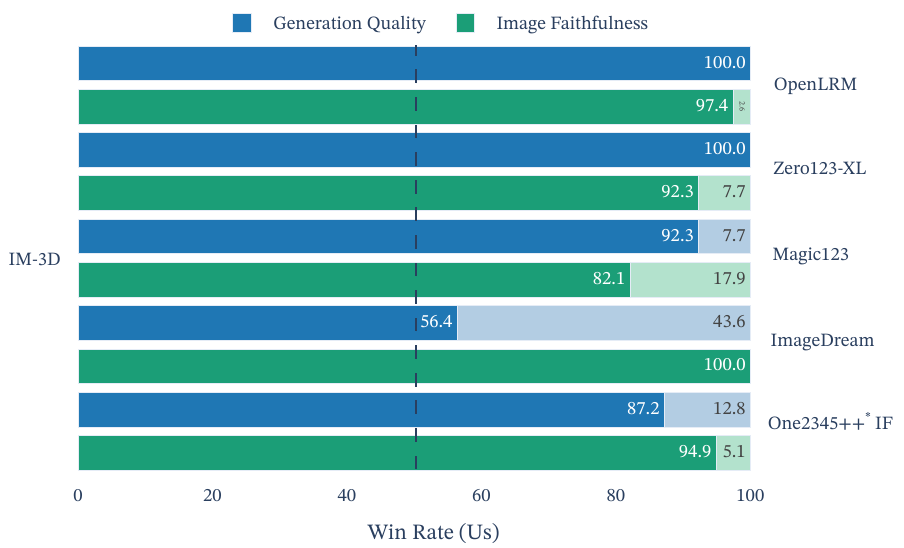}
\vspace{-2mm}
\caption{\textbf{Human evaluation.}
We perform human evaluation of \method vs state-of-the-art in Image-to-3D and Text-to-3D. Human raters preferred \method to all competitors
with regard to both generation quality and faithfulness, often %
by a large margin. 
}%
\vspace{-6mm}
\label{fig:human_evaluation_figure}
\end{figure}

\section{Experiments}%
\label{s:experiments}

Our method generates 3D objects from a textual description~$\p$ and a reference image~$\I$.
In order to compare to prior work, we consider in particular the set of textual prompts from~\cite{shi23mvdream:}, which are often used for evaluation.

Given an input image and prompt $(\I,\p)$, previous methods either \emph{synthesize} a multi-view image sequence $\J$ (usually by means of a generator network), or output a 3D model, or both.
We compare the quality of the produced artifacts visually, utilizing the image sequence $\J$ directly, or corresponding \emph{renders} $\hat \J$ of the 3D model.
In general, we can expect the quality and faithfulness of $\J$ to be better than that of $\hat \J$ because the generated image sequence needs not be perfectly view-consistent.
On the other hand, the renders $\hat\J$ from the 3D model are consistent by construction, but may be blurrier than $\J$, or contain other defects.

\vspace{-2mm}
\subsection{Comparison to the state-of-the-art}

In this section, we compare \method to relevant state-of-the-art methods in the literature, including
MVDream~\cite{shi23mvdream:},
Zero123XL~\cite{deitke23objaverse-xl:},
Magic123~\cite{qian23magic123:},
SyncDreamer~\cite{liu23syncdreamer:},
ImageDream~\cite{wang23imagedream:},
LRM~\cite{hong23lrm:} and
One2345++~\cite{liu23one-2-3-45:}.
For LRM, since no public models are available, we utilize the open-source OpenLRM~\cite{openlrm}. 
For One2345++, which is only available via a web interface, we manually upload each image in the evaluation set. 
We carry out both quantitative and qualitative comparisons using the set of prompts and images from \cite{shi23mvdream:,wang23imagedream:}.

\paragraph{Quantitative comparison.}

\Cref{tab:clip_scores} provides a quantitative comparison of our method to others. %
We adopt the same metrics as~\cite{shi23mvdream:,wang23imagedream:}, which are based on the CLIP~\cite{radford21learning} similarity scores.
Specifically, we utilize the ability of CLIP to embed text and images in the same space.
We then use the embeddings to compare the textual prompt $\p$ and the image prompt $\I$ to the images $\J$ of the object (either synthesized or rendered).
A high CLIP similarity score means high faithfulness to the prompt.
As an upper bound, we also report the CLIP scores of the prompt images $\I$ which were generated using the SDXL~\cite{podell23sdxl:} model.

The key takeaway from \cref{tab:clip_scores} is that \method outperforms all others in terms of both textual and visual faithfulness.
This is true for both the image sequences $\J$ output by the video generator as well as the renders $\hat \J$ from the fitted 3D GS models $G$.
\method is particularly strong when it comes to visual faithfulness, which also means that the images we generate are of a quality comparable to the input image $\I$.
Additionally, our method requires significantly less time than most (3 minutes vs hours for some models).

\paragraph{Human evaluation.}

Automated metrics for the evaluation of generative models are not fully representative of value of the output in applications.
Thus, we also conduct a human study.
We ask annotators to evaluate our model against a competitor based on (1) Image Alignment and (2) 3D quality.
We present to annotators with the outputs of two different methods, rendered as 360\textdegree{} videos, and ask them to indicate a preference based on these two criteria.
\Cref{tab:clip_scores} shows the win rate when comparing our methods against others.
Our method surpasses in performance all other baselines in both studies indicating that the proposed method produces high-quality 3D results that closely align with the image prompt. Further details are provided in the Appendix.

\begin{figure*}
\centering
\includegraphics[width=0.97\textwidth]{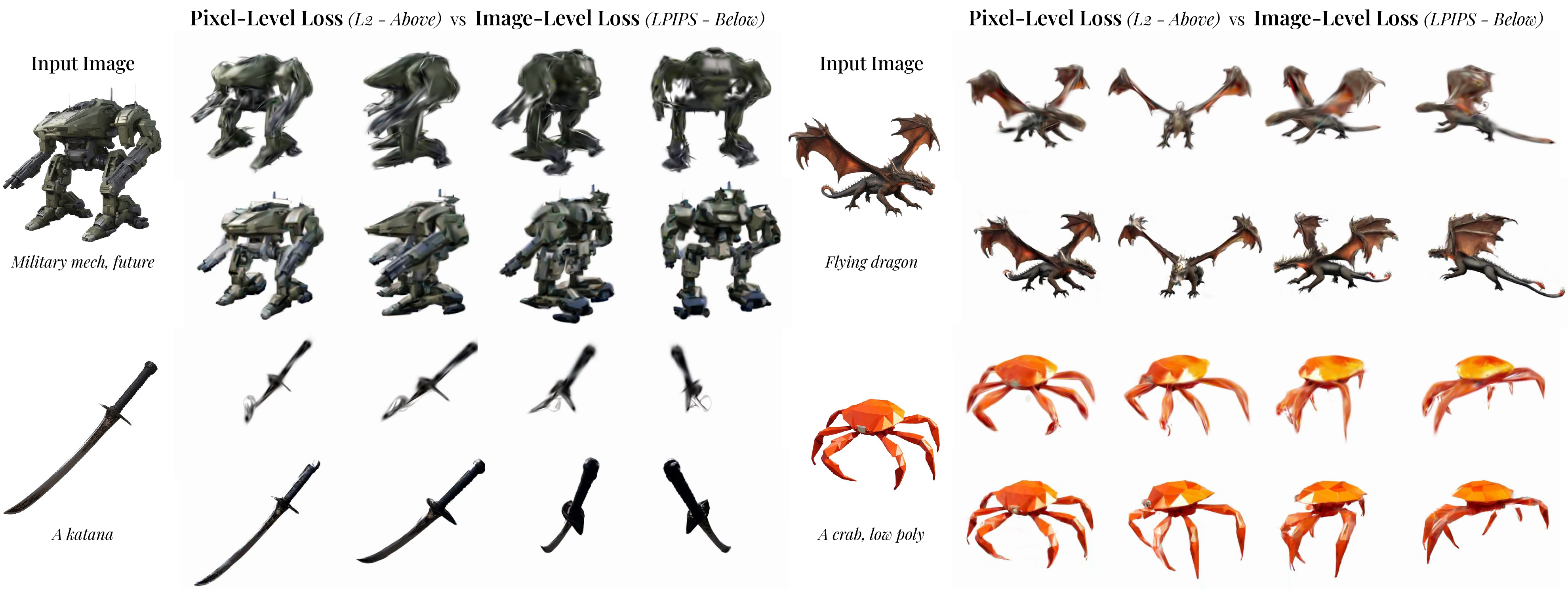}
\caption{\textbf{Reconstruction Quality with Pixel-Level and Image-Level Losses.} 
We find that image-level losses are crucial to the success of our method. 
With pixel-level losses such as the L2 loss, small inconsistencies in the generated images are effectively averaged together, resulting in unnatural and blurry-looking reconstructions.
}%
\label{fig:image_level_loss_figure}
\vspace{-2mm}
\end{figure*}

\subsection{Ablations}
\begin{table}[t]
\small
\centering
\vspace{-2mm}
\caption{\textbf{Ablation on the importance of loss terms and 3D representation during the fitting stage.}}
\label{tab:loss_ablation}
\vspace{-2mm}
\begin{tabular}{lll}
\toprule
Loss / Representation & \multicolumn{1}{c}{\!CLIP (Text)} & \multicolumn{1}{c}{\!\!CLIP (Image)} \\ \midrule
\textbf{\method (ours)} & \textbf{31.66} \textpm{1.7} & \textbf{91.40} \textpm{5.5} \\
- $\mathcal{L}_\text{LPIPS}$ & 29.38 \textpm{2.1} & 84.71 \textpm{6.4} \\
- $\mathcal{L}_\text{RGB}$ instead of $\mathcal{L}_\text{LPIPS}$ & 29.67 \textpm{2.0} & 84.99 \textpm{5.9} \\
- $\mathcal{L}_\text{SSIM}$ & 31.53 \textpm{1.8} & 90.64 \textpm{5.7} \\
- $\mathcal{L}_\text{Mask}$ & 31.43 \textpm{1.9} & 90.14 \textpm{6.0} \\ 
w/ NeRF instead of GS & 30.42 \textpm{2.1} & 87.37 \textpm{5.4} \\ \bottomrule
\end{tabular}
\vspace{-4mm}
\end{table}

\paragraph{Effect of iterative refinements.}

In \cref{fig:iterations_figure}, we demonstrate the efficacy of our proposed iterative refinement process. 
Our model's initial reconstructions (derived from training on our initially generated videos) are compared to the outcome following a single iteration of multiview diffusion and reconstruction.
While the initial reconstructions exhibit satisfactory shapes, they miss out on intricate details due to minor inconsistencies in the initial multiview images.
In a few instances, some parts of these initial reconstructions look as if two copies of a shape have been superimposed upon one another, as the reconstruction process tries to satisfy two inconsistent views. 
However, our technique rectifies these discrepancies with one iteration of denoising and reconstruction; significantly enhancing the level of detail. 

\paragraph{Image-Level Losses.}
In \cref{tab:loss_ablation} and \cref{fig:image_level_loss_figure}, we compare results of optimization with pixel-level and image-level loss functions. 
We find that image-level losses are central to our method's ability to generate high-quality 3D assets. 
The use of pixel-level losses such as L2 loss is detrimental, as minor inconsistencies in the multiview images are emphasized by the optimization process and effectively averaged together. 
This averaging results in a low CLIP score (29.67 vs 31.66 for LPIPS) as well as blurry and unnatural generations. %

\paragraph{Comparing 3D Representations}
The last line of \cref{tab:loss_ablation} provides a comparison of 3D representations, showing the effect of using NeRF as an underlying 3D representation rather than Gaussian splatting (GS). 
We find that the visual quality of models generated using NeRF is slightly worse than GS. 
The true benefit of GS is that it is much faster and much more memory-efficient; training with GS takes 3 minutes whereas training with NeRF takes 40 minutes.
Additionally, the memory-efficient nature of Gaussian splatting makes it easy to render at our diffusion model's native resolution of $512$px, whereas for NeRF one has to use ray microbatching or optimize at a lower resolution. 

\begin{table}[t]
\small
\centering
\caption{\textbf{Ablation on Using Fewer Frames.} We show quantitative performance when performing our reconstruction and generation using fewer frames. }
\begin{tabular}{lll}
\toprule
\# Frames & \multicolumn{1}{c}{CLIP (Text)} & \multicolumn{1}{c}{CLIP (Image)} \\ \midrule
16 & \textbf{31.66} \textpm{1.7} & \textbf{91.40} \textpm{5.5} \\
8 & 31.38 \textpm{1.8} & 90.06 \textpm{6.3} \\
4 & 30.06 \textpm{2.6} & 86.96 \textpm{8.6} \\
\bottomrule
\end{tabular}
\label{tab:skip_frame_scores}
\vspace{-6mm}
\end{table}

\paragraph{Using Fewer Frames} 
Differently from the vast majority of other diffusion-based text-to-3D and image-to-3D approaches, which generate only 1-4 frames, \method generates 16 frames simultaneously. We demonstrate the significance of this in \cref{tab:skip_frame_scores}, finding that our quantitative performance improves as we increase the number of generated frames. 

\subsection{Limitations}%
\label{s:limitations}

The fine-tuned video generator is generally very view-consistent,  but it still has limitations. 
One interesting failure case is that for highly dynamic subjects (e.g., horses, which are often captured running), the model sometimes renders spurious animations (e.g., walking or galloping) despite our fine-tuning, which is problematic for 3D reconstruction. 
This occurs more often when the prompt contains verbs describing motion; see the Appendix for an example. 

\vspace{-2mm}
\section{Conclusions}%
\label{s:conclusions}

In this work, we have shown that starting from a video generator network instead of an image generator can result in better multi-view generation, to a point where it can impact the design of future text-to-3D models.
In fact, we have shown that the quality is sufficient to eschew distillation losses like SDS as well as large reconstruction networks.
Instead, one can simply fit the 3D object to the generated views using a robust image-based loss.
Reconstruction can be further alternated with refining the target video, quickly converging to a better 3D object with minimal impact on efficiency.
Compared to works that rely on SDS, our approach significantly reduces the number of evaluations of the 2D generator network, resulting in a faster and more memory-efficient pipeline without compromising on quality.

\section{Impact Statement} %

Our work uses Generative AI, whose potential impacts are and have been extensively discussed in the academic, business and public spheres.
Our work does not change these issues qualitatively.
The Emu models~\cite{dai23emu:} were explicitly designed with fairness and safety in mind, and fine-tuning them on curated 3D models is likely to further reduce the potential for harm.

\bibliography{references,vedaldi_general,vedaldi_specific}
\bibliographystyle{icml2024}

\appendix
\onecolumn
\section{Appendix}

\subsection{Training details}

In line with~\cite{girdhar23emu-video:}, we maintain the spatial convolutional and attention layers of Emu Video, fine-tuning only the temporal layers. We minimize the standard diffusion loss over a span of 5 days, employing 80 A100 GPUs with a total batch size of 240 and a learning rate of \textit{1e-5}. Our findings indicate that prolonged training effectively counters the network’s inclination to generate 360 videos of deforming objects, given that the initialization is a video generation model. Contrary to MVDream~\cite{shi23mvdream:} and Instant3D~\cite{li23instant3d:}, we observe no degradation in texture quality with extended training. This can be ascribed to the fact that the spatial layers remain static and the network is image-conditioned, necessitating that the generated 360 video retain the high-frequency texture elements of the input.

For Gaussian fitting, we initialize 5000 points at the center of the 3D space, and densify and prune the Gaussians every 50 iterations. We conduct optimization for 1200 iterations and execute Emu Video twice for 10 iterations each using the DPM solver~\cite{lu22dpm-solver:} during this process, repeating this every 500 iterations.  Empirically, we found that setting the weights to  $w_\text{LPIPS}=10$, $w_\text{SSIM}=0.2$ and $w_\text{Mask}=1$ yields the best results during the fitting stage. 

Prior to fitting,  we need to estimate the elevation of the very first generated video. To that end, we trained an elevation estimator on top of DINO~\cite{caron21emerging} features using the 100k 3D renderings. The network averages the features of 4 uniformly distributed frames and uses a 2-layer MLP to regress the elevation in radians.

\subsection{Human evaluation}
In our study, we employed the prompt set delineated in~\cite{shi23mvdream:} to conduct a human annotation evaluation via Amazon Mechanical Turk (AMT). The task assigned to the annotators involved choosing between two 3D assets, both rendered as 360\textdegree{} videos, 
with one of the assets being the output of our proposed method. 
To ensure a robust and unbiased evaluation, we randomized the presentation order of the methods for each question. 
Each question was assessed by five annotators, and we reported the consensus opinion. 
Since each question corresponds to a triplet of (competing method, 3D reconstruction, and quality/faithfulness), this is a total of $5 \text{ annotators} \times (5\text{ methods} \times 39\text{ reconstructions} \times 2\text{ question types}) = 1950$ annotations.
We instructed the annotators to overlook any disparities in background colors, as normalizing all methods to yield the same scale is a non-trivial task.

\subsection{Further information on network efficiency}%

\begin{table}[h!]
\centering
\begin{tabular}{lcccccc}
\toprule
 & ProlificDreamer & MVDream & ImageDream & Zero123XL & SyncDreamer & \method \\ \midrule
\begin{tabular}[c]{@{}l@{}}Number of \\ network calls\end{tabular} & 320000 & 20000 & 25000 & 1200 & 200 & 80 \\ \bottomrule
\end{tabular}
\caption{\textbf{Number of diffusion network calls to generate one 3D asset.} The proposed method, \method, requires only a fraction of the model evaluations to compute a 3D asset.}
\label{tab:num_of_network_calls}
\end{table}

In \cref{tab:num_of_network_calls}, we present the number of diffusion model forward passes used to reconstruct a single 3D object for various 3D generation methods. Whereas some other methods require thousands or tens of thousands of iterations, our method requires less than one hundred. 

\subsection{Failure Cases}%
\label{s:failure_cases}
\begin{figure*}
\centering
\includegraphics[width=0.57\textwidth]{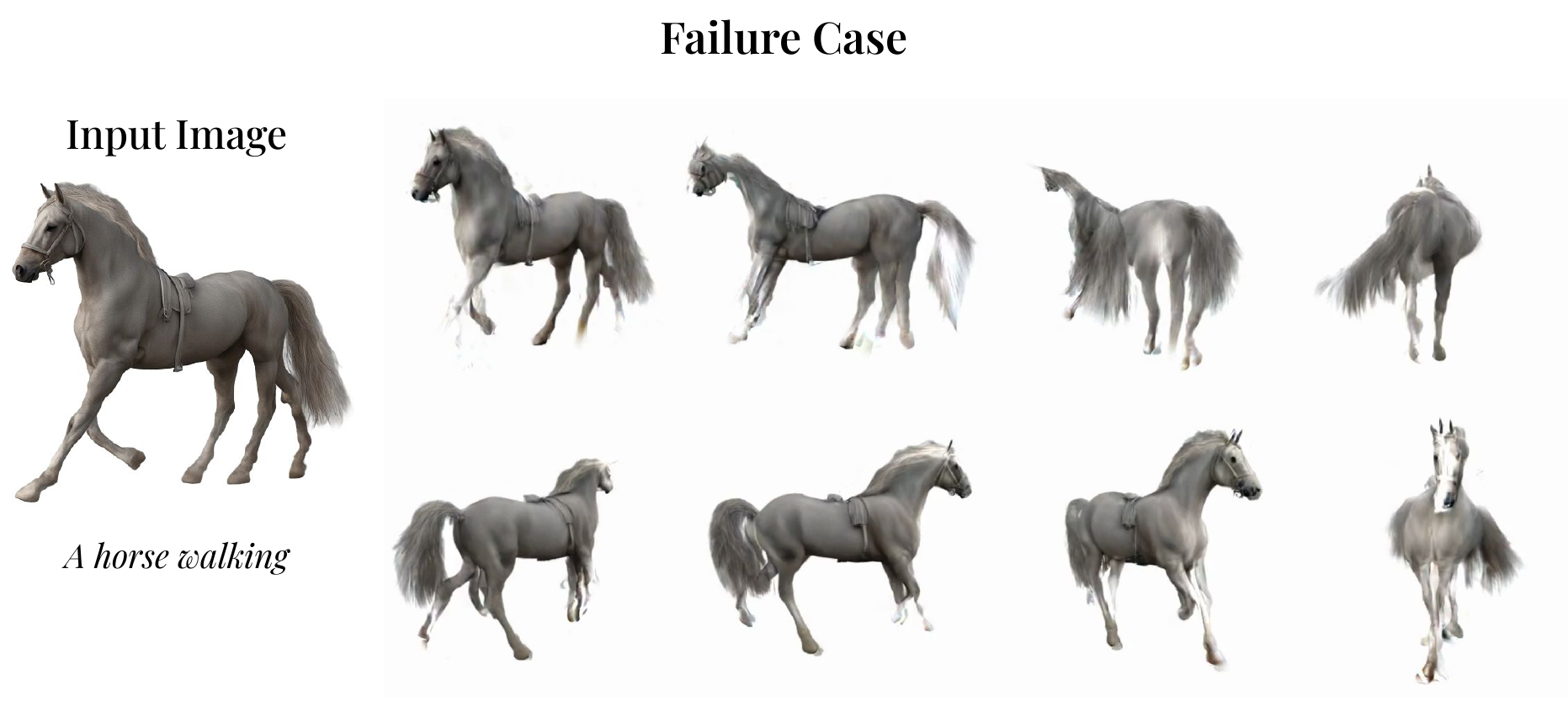}
\caption{\textbf{Visualisation of a failure case.} 
In this case, our finetuned video network generated an animated video of a horse walking rather than a static video. As a result, the 3D reconstruction process produces erroneous geometry (e.g. the head of the horse is barely visible from some views). 
}%
\label{fig:failure_figure}
\end{figure*}

As described in the limitations section of the main paper, the video generator does not produce perfect results in all cases. 
A notable instance of failure is observed with subjects that exhibit high dynamism, such as horses often depicted in motion. In such cases, the model occasionally produces unwarranted animations like walking or galloping, which disrupts the 3D reconstruction process. 
We show an example in \cref{fig:failure_figure}. 

\subsection{Conversion to Meshes}%
\label{s:meshes}

\begin{figure*}
\centering
\includegraphics[width=0.97\textwidth]{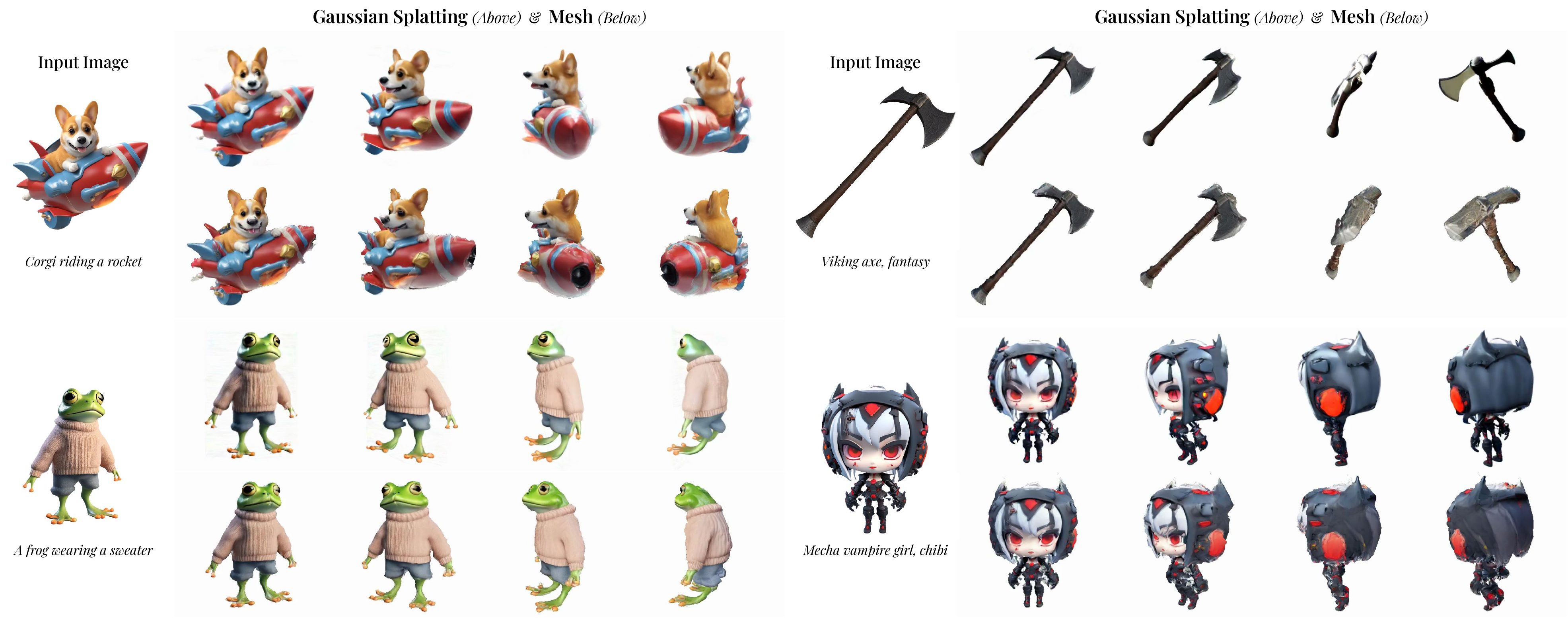}
\caption{\textbf{Visualisation of Meshes.} 
We convert our Gaussian Splatting representation to DMTet~\cite{shen2021deep} using marching cubes and optimize the resulting meshes using our iterative multiview diffusion and reconstruction process.
}%
\label{fig:meshes_figure}
\end{figure*}

Although Gaussian Splatting is being rapidly adopted by the computer vision and graphics communities, some production applications require that objects be converted to meshes. We show that in these cases, it is straightforward to extract high-quality meshes from our Gaussian Splatting representation: one can run marching cubes~\cite{cline88admissibile} and then optionally continue to optimize the resulting mesh using DMTet~\cite{shen2021deep}. We show visual results of converting our models to meshes in \cref{fig:meshes_figure}.

\end{document}